%% file: root.tex
\documentclass{article}
\usepackage{float} 
\usepackage{cite}
\usepackage[margin=0.5in]{geometry} 
\usepackage{amsmath,amssymb,amsfonts}
\usepackage{algorithmic}
\usepackage{graphicx}
\usepackage{textcomp}
\usepackage{xcolor}
\usepackage{url}

\newcommand{\ourAcronym}{D2HDMap}

\usepackage{tikz}
\newcommand{\ieeeNotice}[1]{%
  \begin{tikzpicture}[remember picture,overlay]
    \node[anchor=south, yshift=3cm] at (current page.south) {%
      \def\arraystretch{1.2}
      \begin{tabular}{|p{\dimexpr\textwidth-2\tabcolsep-2\arrayrulewidth\relax}|}
        \hline
        \small #1 \\
        \hline
      \end{tabular}
    };
  \end{tikzpicture}%
}

\def\BibTeX{{\rm B\kern-.05em{\sc i\kern-.025em b}\kern-.08em
    T\kern-.1667em\lower.7ex\hbox{E}\kern-.125emX}}

\begin{document}

\title{\ourAcronym: Non-visible Driveline Map Prior for \\Online Vectorized HD Map Prediction

\thanks{This work done during Seojun Shon's internship at Nissan North America.}
}

\author{Seojun Shon$^{1}$, Chikao Tsuchiya$^{1}$, Dhaval Bhanderi$^{1}$,\\David Ilstrup$^{1}$, Hsinmin Cheng$^{1}$ and Christopher Ostafew$^{1}$
\thanks{$^{1}$Chikao Tsuchiya, Dhaval Bhanderi, David Ilstrup, Hsinmin Chen, and Christoper Ostafew are with Nissan Advanced Technology Center - Silicon Valley, Nissan North America, Santa Clara, CA 95014, USA
        {\tt\small \{Chikao.Tsuchiya, Dhaval Bhanderi, David.Ilstrup, Hsinmin.Cheng, Christopher.Ostafew\}@nissan-usa.com}}%
}

\maketitle
{\ieeeNotice{© 2026 IEEE. Personal use of this material is permitted. Permission from IEEE must be obtained for all other uses, in any current or future media, including reprinting/republishing this material for advertising or promotional purposes, creating new collective works, for resale or redistribution to servers or lists, or reuse of any copyrighted component of this work in other works. To appear in: 2026 Intelligent Vehicles Symposium (IV), Proceedings}
}


\input{0_abstract}

\input{1_intro}
\input{2_relatedWork}
\input{3_method}

\input{4_experiments}
\input{5_conclusion}

\bibliographystyle{unsrt} 
\bibliography{mapping}

\end{document}

%% file: 0_abstract.tex
\begin{abstract}

Accurate, up-to-date representations of road structures are critical for the safe operation of autonomous vehicles.
Existing systems rely either on costly, maintenance-heavy high-definition (HD) maps—which compromise safety when outdated—or purely sensor-based online mapping, which struggles with long-range reliability and occlusion.
Systems incorporating map prior information into online mapping seek to overcome drawbacks of both approaches by combining them in some way.
We propose \textit{`Driveline To HD Map'} \textit{(\ourAcronym)} -- an online mapping system that injects a lightweight, non-visible driveline prior to guide the estimation of visible road structures such as lane dividers, road boundaries and crosswalks. This prior incurs less effort to create and update compared to full HD map priors used in other approaches. We also show that training with such a prior can improve generalization at inference time when no prior is available. Ablation studies conducted on the nuScenes and Argoverse 2 dataset demonstrate that models trained using a driveline prior retain performance even when priors are not available.
On a geographically disjoint split, \ourAcronym{} achieves \textbf{44.8} mAP, surpassing recent state-of-the-art. Additionally, noise-aware training substantially increases robustness to realistic localization error.


\end{abstract}

%% file: 1_intro.tex
\section{Introduction}

Autonomous vehicle safety is fundamentally dependent on a precise and up-to-date road structure representation. HD maps are well established as a mechanism to provide this information for safe navigation and prediction \cite{elghazaly2023high}. While effective, the cost of creating and maintaining HD maps, coupled with drawbacks such as the need to update the map when the roadway is modified, or the misleading nature of the map in the presence of construction, temporary lanes, etc., present significant challenges that must be addressed.
  
Interest in online mapping has increased recently as an answer to many of these drawbacks. Onboard sensors provide the information to estimate critical map elements on-the-fly, reflecting current road conditions. Estimates are made in the car frame, avoiding an extra localization step to register map elements. While online mapping systems do work around some HD map issues, they introduce problems of their own. Inherent limitations of onboard sensing make it difficult to estimate map structures at distances required for safe prediction and planning, under occlusion, and in visually challenging conditions such as night time or adverse weather. 

Online systems that incorporate map prior information seek to leverage some or all of the strengths of HD maps while continuing to overcome their weaknesses. The prior representations employed in the literature vary considerably, motivated by both what is available and by the goal of including the prior.

\newpage
\begin{figure}[t]
    \centering
    \includegraphics[width=0.45\textwidth]{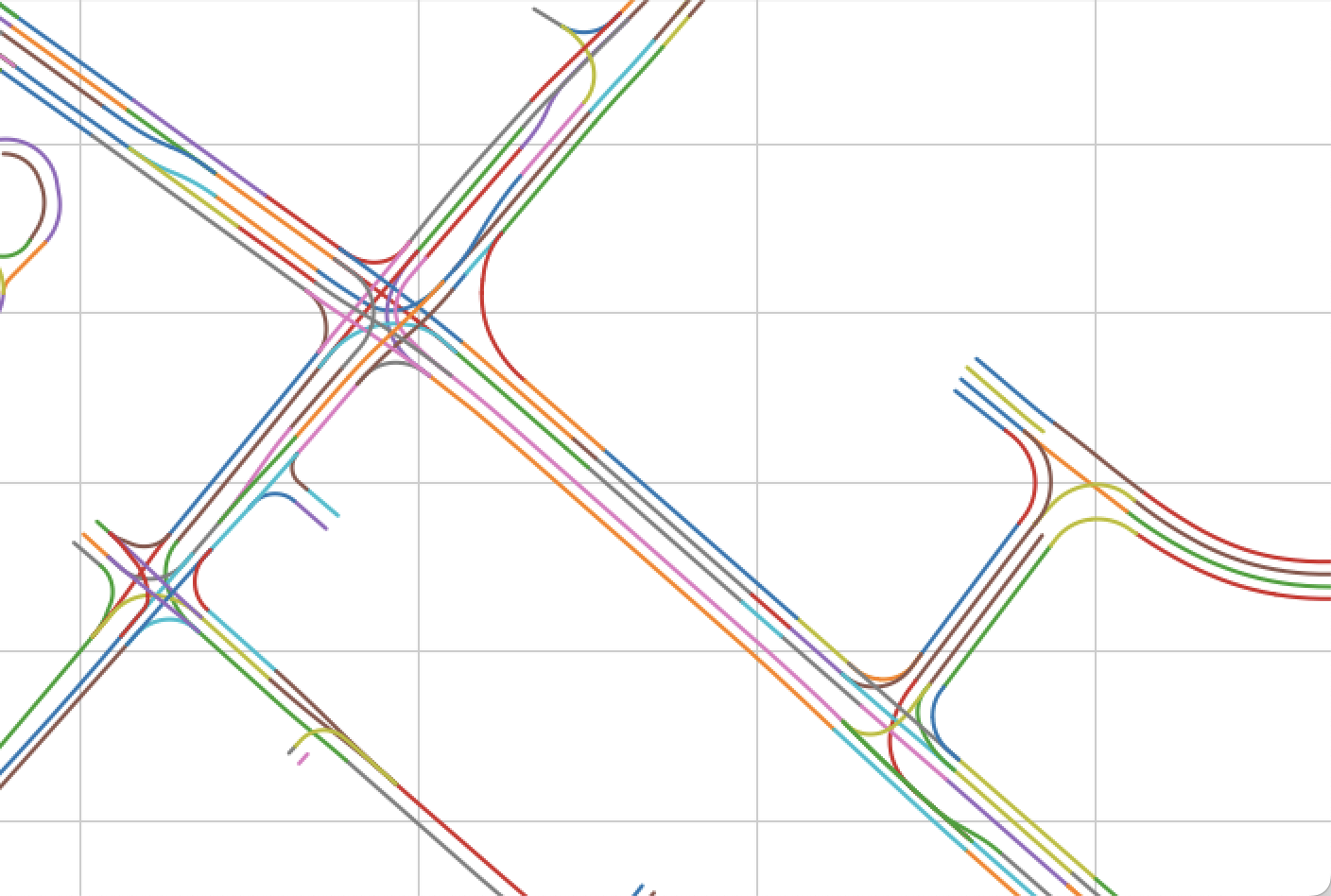}   
    \caption{Illustration of a driveline map prior. Each solid curve represents a lane’s driveline—serving as a lightweight, non-visible geometry cue that guides the prediction of visible elements such as lane dividers, road boundaries, and crosswalks.}
    \label{fig:drivelinemap}
\end{figure}

We propose an online mapping system dubbed \textit{`Driveline To HD Map' (\ourAcronym{})} that builds on a previous system by replacing visible map element priors with a lightweight driveline map prior shown in Fig.~\ref{fig:drivelinemap}. This provides a strong geometry cue for the lanes without predefining the exact geometry of the visible markings. We give evidence that this preserves flexibility when priors are sparse or missing and reduces shortcut learning. To handle real-world localization error, we inject pose perturbations during training, aiding model robustness at inference time. 

In summary, our main contributions are:
\begin{itemize}
    \item Propose driveline only priors as support for inference of more diverse visible map elements (\ourAcronym).
    \item Show improved robustness by expanding simulated localization noise during training.
    \item Demonstrate generalization and robustness of the method with ablation studies, geographic splits, and simulated localization noise.
\end{itemize}

This work aims to provide a scalable, cost-effective, and informative map prior while retaining its critical functionality for autonomous driving systems.

%% file: 2_relatedWork.tex
\section{Related Work}

We focus on recent work in the style of MapTR \cite{liao2023maptr} using camera input. A typical implementation begins with an image backbone such as ResNet \cite{He_2016_CVPR} or DINOv2 \cite{oquab2023dinov2} 
and produces vector or poly-line representations of estimated map elements, such as lane dividers.

\subsection{Rasterized output to vectorized output}

Recent trends favor vectorized HD-map elements rather than rasterized grids because rasterized maps lack structural information such as direction, keypoints and instance IDs and they struggle to represent complex topologies or over-lapping elements. These drawbacks make them less suitable for downstream tasks such as prediction and planning. 

An early work, HDMapNet \cite{li2022hdmapnet} produces rasterized elements requiring post-processing to extract vector representations while VectorMapNet \cite{liu2023vectormapnet} learns to decode poly-lines directly in an end-to-end fashion. MapTR \cite{liao2023maptr} produces vectorized output, introducing innovations such as training with vertex permutations to make element identification more robust and MapTRv2 \cite{liao2025maptrv2} with decoupled attention to better use processing time. Subsequent advances improve temporal stability \cite{yuan2024streammapnet, chen2024maptracker, yang2025histrackmap} and explore generative formulations \cite{monninger2025mapdiffusion} to better handle sparsity and occlusions.

\subsection{Priors}

Priors used by works we consider fall into several forms: temporal prior, map prior, and others. Temporal consistency and tracking  approaches such as StreamMapNet \cite{yuan2024streammapnet}, MapTracker \cite{chen2024maptracker}, and HisTrackMap \cite{yang2025histrackmap} progressively improve map consistency and accuracy through query propagation and historical memory techniques. HRMapNet \cite{zhang2024enhancingvectorized}, PrevPredMap \cite{peng2025prevpredmap} and its successor, Uni-PrevPredMap \cite{peng2025uni} employ a temporal prior that is updated with the predictions of each frame.

Map priors are drawn from an external representation of map information in some coordinate system. Localization is used to select an appropriate subset of this information and register it in the car frame. SMERF \cite{luo2023augmentinglaneperceptiontopology} and P-MapNet \cite{jiang2024p} utilize SD maps \cite{OpenStreetMap} with P-Map also including HD-map prior information. NMP \cite{xiong2023neural} utilizes past detection results by maintaining a neural representation of global maps that is incrementally updated with features from previous traversals. NeMO \cite{zhu2023nemo} leverages a readable and writable big feature map that stores historical BEV features, acting as a repository for past perceptions. MapEX \cite{sun2025mind}, LDMapNet-U \cite{xia2025ldmapnet}, Uni-PrevPredMap \cite{peng2025uni}, and Nuro's method \cite{Bateman2024hdmapprior} rely on integrating existing, historical, or simulated outdated HD maps as priors for map generation and updates. 

SD++ \cite{diwanji2025sd++}, \cite{tumu2025usinglanguageroadmanuals} and SDTagNet \cite{immel2025sdtagnet} incorporate textual annotations or LLM enhancements as a prior. Additionally, TrajTopo \cite{jia2025enhancing} and TrailTR \cite{hubbertz2025inferring} explore the use of trajectory or trail data, combined with SD maps in the case of TrajTopo, to capture dynamic road structures and driving patterns, while SMART \cite{ye2025smart} uniquely combines SD maps and satellite images with supervision from large-scale HD maps for scalable topology reasoning.

\ourAcronym{} incorporates both temporal and map priors. 
We leverage historical predictions and a non-visible driveline map as sources of prior information to infer the visible elements. 

\subsection{Localization Error}

Localization error affects the registration of map prior data to the car frame, and so must be considered once such priors are introduced. Even small translation or yaw errors can mis-register priors and degrade online mapping performance significantly. In \cite{Reid_2019} the authors reason about acceptable localization errors using models of car and roadway dimension and vehicle speed. Some existing studies \cite{zhang2024enhancingvectorized, Bateman2024hdmapprior, sun2025mind, zhang2024ucsd} highlight how ego translation and rotation errors directly impact accuracy of map estimation. For evaluation, these errors are simulated by adding random noise to poses taken as ground truth, typically following Gaussian distributions. This noise affects various processes, including priors retrieval and updating. It has been demonstrated that the performance substantially deteriorates as pose noise increases. 
However, many works \cite{yuan2024streammapnet, chen2024maptracker, peng2025prevpredmap, luo2023augmentinglaneperceptiontopology, jiang2024p, xiong2023neural, zhu2023nemo, xia2025ldmapnet, jia2025enhancing, hubbertz2025inferring, ye2025smart} do \emph{not} explicitly harden models using training data augmented with localization noise, while others \cite{zhang2024enhancingvectorized, peng2025uni, sun2025mind} are trained with unrealistically small localization noise, using only small translation error, but no rotation error.

 
 \ourAcronym{} adopts simple, deployment-friendly noise-aware training and explores how augmenting training with various levels of translation and rotation errors affects inference performance both with and without simulated errors.

%% file: 3_method.tex
\section{Methodology}

\subsection{Overall Architecture}

\begin{figure*}[ht]
    \centering
    \includegraphics[width=1.0\textwidth]{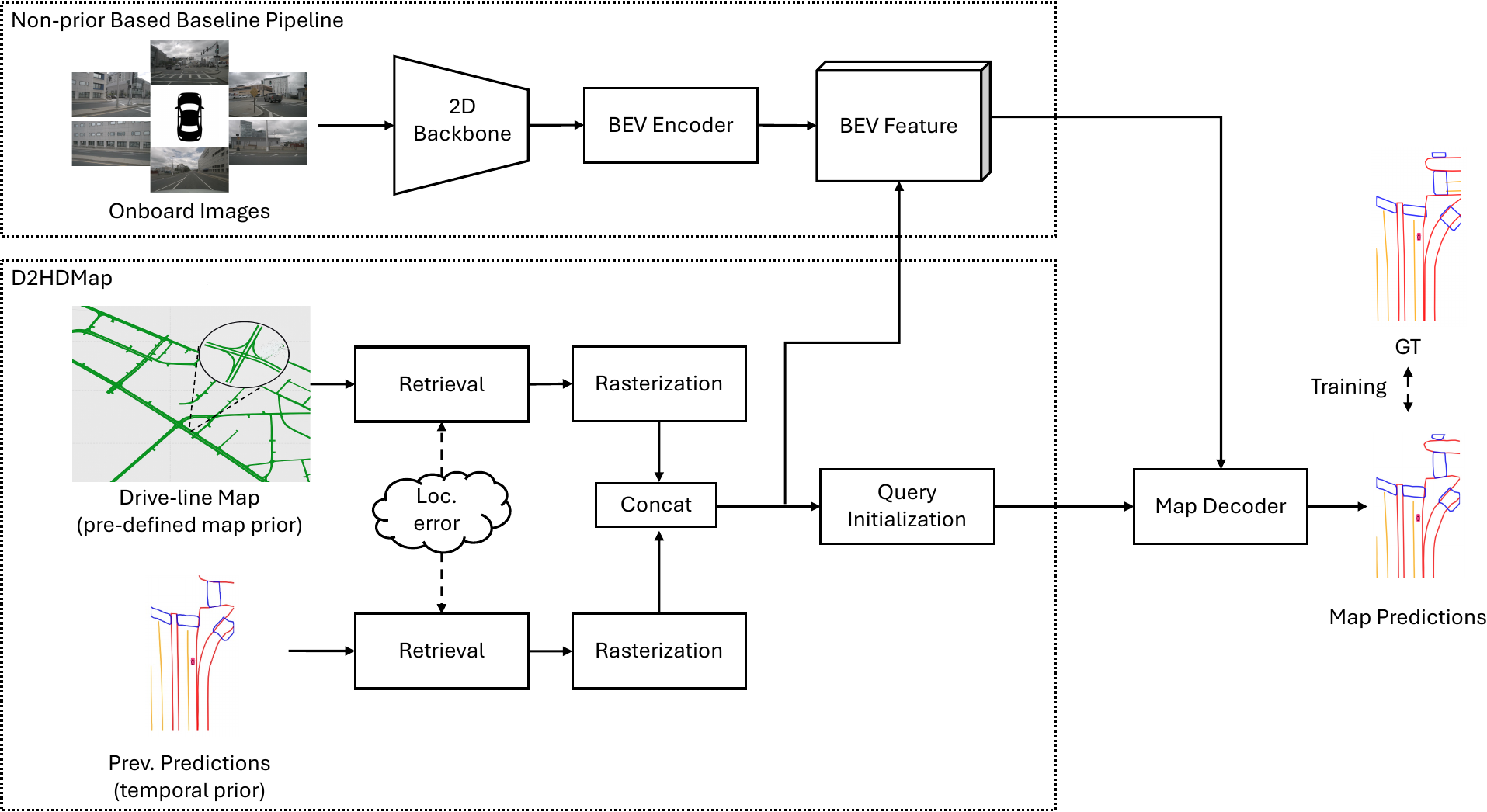}   
    \caption{System diagram of \ourAcronym. Our proposed architecture builds on the Uni-PrevPredMap framework by incorporating a driveline-based prior to guide online vectorized map construction. The driveline prior is encoded as a structured sequence and fused with map queries via cross-attention, allowing the network to align predicted map elements with these motion-feasible trajectories.}
    \label{fig:system_diagram}
\end{figure*}

\ourAcronym{} is built on the Uni-PrevPredMap \cite{peng2025uni} framework by incorporating a driveline-based prior to guide online vectorized map construction. In our experiments, the prior is derived from nuscenes \cite{caesar2020nuscenes} and argoverse 2 \cite{wilson2023argoverse} centerlines. 
Following the treatment of map priors in \cite{peng2025uni}, the driveline prior is rasterized and along with the rasterized temporal prior influences downstream query and feature formation.

As illustrated in Fig.~\ref{fig:system_diagram}, \ourAcronym{} captures surrounding images using multiple cameras mounted on the vehicle. These images are processed through a perspective-view (PV) backbone \cite{He_2016_CVPR} to extract 2D features, which are then transformed into BEV features \cite{huang2022bevpoolv2}. Furthermore, \ourAcronym{} incorporates two types of vectorized maps as prior information: the \textbf{driveline map prior} and the \textbf{temporal map prior}, enhancing the accuracy and robustness of the map estimation process. To retrieve prior information, \ourAcronym{} accesses the driveline and temporal maps based on the current ego pose. These maps are rasterized and concatenated to form a unified prior representation, which is then used to initialize the map queries following the same strategy as Uni-PrevPredMap \cite{peng2025uni}. The accuracy of the ego vehicle’s pose directly affects the quality of query initialization, as it determines the spatial alignment of the retrieved priors.

During decoding, the map decoder performs cross-attentions between the prior-informed queries and the BEV features to predict map elements. The predicted elements are transformed into the global coordinate frame using the ego vehicle’s pose and are added to the temporal map. Again, the precision of the ego pose is critical at this stage, as it governs the correct placement and consistency of map elements over time.

\subsection{Map Priors}

Prior information improves spatial consistency and semantic accuracy in online map estimation. Priors such as pre-defined maps and temporally accumulated detection maps provide structured contextual cues that guide the network toward plausible predictions, particularly under occlusions, sparse observations, or, in the case of pre-defined maps, adverse conditions such as nighttime or heavy rain. Retrieval and alignment of these priors based on the ego vehicle’s pose ensure spatially grounded query initialization, which directly affects downstream prediction quality.
Pre-defined map priors add significant information at ranges where sensors struggle.


A key contribution of our framework is the integration of a driveline-based prior, which introduces structured motion constraints into the estimation process. Unlike conventional priors relying solely on past detections or static elements, the driveline prior encodes vehicles’ intended trajectories as a structured sequence, providing predictive guidance even in visually degraded or occluded regions. This enables the model to infer motion-feasible map elements aligned with plausible future paths.

Critically, both the retrieval of the driveline prior and the global alignment of predicted map elements depend on accurate ego pose estimation. Any pose error directly impacts query initialization and temporal map updates. By tightly coupling the driveline prior with pose-aware query initialization, our method improves robustness and semantic coherence in challenging and partially observable environments.

\subsection{Training with Localization Errors}
While localization technologies in autonomous driving continue to advance, localization errors remain inevitable and can significantly affect online map estimation with map priors. Uni-PrevPredMap \cite{peng2025uni} addresses this issue by introducing translation errors in their outdated HD map simulation, capturing effects such as missing or shifted map elements. However, it does not account for rotation errors, which can severely impact the alignment of distant map elements.

To reflect realistic conditions, our framework considers both translation and rotation errors during training, enabling the model to learn robustness against pose inaccuracies. Both translation and rotation errors are drawn from a Gaussian distribution with 0 mean and pre-defined $\sigma$. As described in Tab.~\ref{table:localization_error}, we considered 4 different values of $\sigma$ for translation and 3 for rotation. Furthermore, when predicted map elements are added to the temporal map, they are subject to the same localization errors as during prior retrieval. To faithfully simulate this, we use the same perturbed pose for both retrieval and accumulation, ensuring consistent treatment of localization uncertainty throughout the pipeline.

%% file: 4_experiments.tex
\section{Experiments}

\subsection{Datasets and Evaluation Metrics}
We evaluate our online mapping framework primarily on the nuScenes dataset \cite{caesar2020nuscenes} to assess generalization across diverse urban environments.

nuScenes is a large-scale autonomous driving dataset comprising 1,000 scenes collected in Boston and Singapore, each lasting 20 seconds and captured with a full sensor suite including 6 cameras. It provides high-definition semantic maps containing lane geometries and pedestrian crossings. Its rich sensor data and temporal continuity make it well-suited for evaluating online mapping algorithms that rely on sensor fusion, and incremental map updates.

Recent studies have identified significant geographical overlap in the original train/validation/test splits of nuScenes, which can lead to inflated performance due to implicit memorization of map features \cite{yuan2024streammapnet, lilja2024localization}. To address this, we adopt the geographically-disjoint splits (geosplit) proposed in SteamMapNet \cite{yuan2024streammapnet}, ensuring that evaluation samples are spatially separated from training data. This setup better reflects the generalization capability of online mapping models to unseen environments and avoids data leakage.

Argoverse 2 complements our evaluation by offering a broader geographic and environmental scope. It includes data from six U.S. cities, with annotated 3D cuboids for 30 object classes and HD vector maps detailing lane topology, traffic direction, and ground height. 

We trained 24 and 6 epochs for nuScenes and Argoverse 2 respectively to compare \ourAcronym{} with others. For evaluation, we adopt Average Precision (AP) as the primary metric, using Chamfer distance as the matching criterion. AP is computed across three spatial thresholds: \{0.5m, 1.0m, 1.5m\}, and the final mean Average Precision (mAP) is obtained by averaging over three road element types: \textit{pedestrian crossing}, \textit{lane-divider}, and \textit{road-boundary}. We use geosplit data for these evaluations.

\subsection{Main Result}

\begin{table*}[ht]
\centering
    \caption{Comparisons with state-of-the-art methods at perception range 60 m × 30 m on nuScenes with geosplit \cite{yuan2024streammapnet}. The results for methods other than \ourAcronym{}  were cited from \cite{monninger2025mapdiffusion}. \ourAcronym{} outperforms in divider, boundary, and overall categorical mAP.}
    \label{table:comparison_with_sota}
    \begin{tabular}{c|c|c|c|c}
        \hline
        Methods & $AP_{div}$ & $AP_{ped}$ & $AP_{bound}$ & $mAP$ \\
        \hline
        VectorMapNet & 17.0 & 15.8 & 21.2 & 18.0 \\
        MapTR & 23.0 & 7.5 & 35.8 & 22.1 \\
        MapVR & 22.6 & 10.1 & 35.7 & 22.8 \\
        MGMap & 25.6 & 7.9 & 37.4 & 23.7 \\
        MapTRv2 & 28.7 & 16.2 & 44.8 & 29.9 \\
        SQD-MapNet & 27.4 & 31.6 & 40.4 & 33.1 \\
        StreamMapNet & 27.3 & 31.2 & 42.8 & 33.8 \\
        MapDiffusion & 31.4 & \textbf{32.9} & 42.4 & 35.6 \\
        \ourAcronym & \textbf{48.6} & 22.4 & \textbf{63.3} & \textbf{44.8} \\
        \hline
\end{tabular}
\end{table*}

In Tab.~\ref{table:comparison_with_sota}, \ourAcronym{} achieves the state-of-the-art performance, \textbf{44.8 mAP}, outperforming the previous best, MapDiffusion \cite{monninger2025mapdiffusion} by 9.2 points on geographically disjoint nuScenes split (60\,\text{m} $\times$ 30\,\text{m} range) proposed by StreamMapNet \cite{yuan2024streammapnet}. The gains are focused on highly represented classes such as lane dividers 48.6 and road boundaries 63.3 which align with the inductive bias of our driveline prior. Since drive-lines encode lane-centric topology and continuity, they provide strong spatial cues for long, lane-aligned structures, helping the model predict dividers and boundaries well even under occlusion or adverse weather.

Performance on pedestrian crossings is lower (22.4 vs. 32.9). This is expected: drive-lines provide little direct signal for crosswalks, which are typically orthogonal to lane flow and governed by intersection semantics rather than lane geometry. In other words, the driveline map prior is highly informative for lane-aligned features, but weakly informative for crosswalks where stop-lines, for example, could be expected to be a better signal.

\subsection{Ablation Study}

\begin{table*}[h]
    \centering
    \caption{Ablation on different priors from nuScenes with geosplit. HD map includes lane divider, pedestrian crossings, and road boundary. HD + Driveline map adds drive-lines on top of HD map. Driveline only map consists of solely drive-lines. $AP^{-}$ means $AP$ without prior in test time.}
    \label{table:different_priors}
    \begin{tabular}{c|c|c|c|c|c|c|c|c|c}
        \hline
        Map Prior & $AP_{div}$ & $AP_{ped}$ & $AP_{bound}$ & $mAP$ & $AP_{div}^{-}$ & $AP_{ped}^{-}$ & $AP_{bound}^{-}$ & $mAP^{-}$ & $\Delta$ $mAP$ \\
        \hline
        HD & 92.9 & 63.9 & 85.2 & 80.7 & 36.4 & 23.7 & 55.1 & 38.4 & -42.3 \\
        HD + Driveline & 94.1 & 73.7 & 94.9 & \textbf{87.6} & 36.3 & 25.9 & 55.1 & \textbf{39.1} & -48.5 \\
        Driveline only & 48.6 & 22.4 & 63.3 & 44.8 & 35.4 & 22.9 & 55.1 & 37.8 & \textbf{-7.0} \\
        \hline
    \end{tabular}
 \end{table*}
 
\begin{table*}[h]
    \centering
    \caption{Ablation on different priors from Argoverse 2 with geographically-disjoint split. HD map includes lane divider, pedestrian crossings, and road boundary. HD + Driveline map adds drive-lines on top of HD map. Driveline only map consists of solely drive-lines. $AP^{-}$ means $AP$ without prior in test time.}
    \label{table:different_priors_av2}
    \begin{tabular}{c|c|c|c|c|c|c|c|c|c}
        \hline
        Map Prior & $AP_{div}$ & $AP_{ped}$ & $AP_{bound}$ & $mAP$ & $AP_{div}^{-}$ & $AP_{ped}^{-}$ & $AP_{bound}^{-}$ & $mAP^{-}$ & $\Delta$ $mAP$ \\
        \hline
        HD & 96.9 & 95.6 & 96.1 & \textbf{96.2} & 53.6 & 54.9 & 66.6 & \textbf{58.4} & -37.8 \\
        HD + Driveline & 96.4 & 93.8 & 96.0 & 95.4 & 50.7 & 48.9 & 62.5 & 54.1 & -41.3 \\
        Driveline only & 68.8 & 53.9 & 74.4 & 65.7 & 51.8 & 51.5 & 63.3 & 55.5 & \textbf{-10.2} \\        \hline
    \end{tabular}
\end{table*}

In Tab.~\ref{table:different_priors} and Tab.~\ref{table:different_priors_av2}, the HD map refers to the original Uni-PrevPredMap model. The Driveline map corresponds to our proposed \ourAcronym{}
representing a lower bound for performance. In contrast, HD + Driveline map denotes \ourAcronym{} equipped with prior information on all available map elements, serving as the upper bound. This comparison illustrates the impact of the driveline map prior on mapping accuracy and highlights the performance envelope of our method.

Interestingly, our method exhibits strong robustness even when the prior information is removed at inference time. Specifically, the model trained only with the driveline map prior performs competitively with the models trained with the more comprehensive map priors in this case, and the absolute drop in performance when the priors are removed is small compared to the models with richer priors.


For the nuScenes geosplit Tab.~\ref{table:different_priors_av2}, \ourAcronym{} performs at SOTA levels (mAP 44.8), despite being out performed by the models using more exhaustive priors. Since this experiment makes no attempt to introduce map changes or localization error, the high mAP values for the HD (80.7) and HD+Driveline (87.6) cases probably reflect high reliance on the prior which is essentially ground truth in this case.
When the priors are removed, mAP's drop to similar levels for all 3 models, with 37.8 mAP for the Driveline trained model and 39.1 mAP, only 1.3 more, for the HD+Driveline trained model.
This is still SOTA performance and the performance drop when removing the prior for the driveline only method is -7.0 mAP.

For the Argoverse 2 geosplit Tab.~\ref{table:different_priors_av2}, results are closer with 57.0 mAP for the Driveline model and 64.3 mAP for the full HD+Driveline model. Similar comments apply as in the nuscenes case. Here the difference in performance between models in the prior-free experiments is 1.4 mAP and the difference in performance for the Driveline only model between using the prior and the prior-free case is -1.6 mAP.

This suggests that the model trained with the driveline prior exhibits competitive generalization, even without relying on explicit priors during inference. HD-map priors overlap the prediction targets (dividers, crossings, boundaries), which likely encourages shortcut learning by memorizing the answer key rather than learning to infer it (seen in the large performance drops between tests with and without the prior).


\subsection{Localization Error}
To assess the robustness of \ourAcronym{} to localization errors, we inject controlled perturbations in both translation and rotation generated pseudo-randomly using a Gaussian distribution.


\begin{table}[h]
\centering
\caption{Performance with various localization errors. Values inside parentheses represent performance drop in percentage from the baseline where no localization error is introduced.}
\label{table:localization_error}
\begin{tabular}{c|ccc}
\hline
Translation & \multicolumn{3}{c}{Rotation Error} \\
 Error & $0.0^{\circ}$ & $2.5^{\circ}$ & $5.0^{\circ}$ \\

\hline
0.0m & 44.8 (baseline) & 37.3 (16.7\%) & 34.7 (22.5\%) \\
0.5m & 38.8 (13.4\%) & 35.1 (21.7\%) & 34.3 (23.4\%) \\
1.0m & 36.0 (19.6\%) & 33.3 (25.7\%) & 33.9 (24.3\%) \\
2.0m & 31.8 (29.0\%) & 31.7 (29.2\%) & 32.3 (27.9\%) \\
\hline
\end{tabular}
\end{table}

\begin{table}[H] 
    \centering
    \caption{Comparison of robust variants with the Non-robust on nuScenes with its geosplit. Robust indicates models trained with localization error while non-robust indicates models trained without localization error. *Robust and Non-robust variants result in the same mAP when no localization error is injected.}
    \label{table:comparison_with_non-robust}
    \begin{tabular}{c|c|c|c|c}
        \hline
        Translation Error & Rotation Error & $mAP$ [Robust] & $mAP$ [Non-robust] & $\Delta$ $mAP$ \\
        \hline
        0.0m & $0.0^{\circ}$ & 44.8* & 44.8* & 0.0\\
        0.0m & $2.5^{\circ}$ & 37.3 & 35.9 & 1.4\\
        0.0m & $5.0^{\circ}$ & 34.7 & 31.2 & 3.5\\
        0.5m & $0.0^{\circ}$ & 38.8 & 39.1 & -0.3\\
        0.5m & $2.5^{\circ}$ & 35.1 & 33.3 & 1.8\\
        0.5m & $5.0^{\circ}$ & 34.3 & 30.1 & 4.2\\
        1.0m & $0.0^{\circ}$ & 36.0 & 32.2 & 3.8\\
        1.0m & $2.5^{\circ}$ & 33.3 & 29.8 & 3.5\\
        1.0m & $5.0^{\circ}$ & 33.9 & 27.8 & 6.1\\
        2.0m & $0.0^{\circ}$ & 31.8 & 26.0 & 5.8\\
        2.0m & $2.5^{\circ}$ & 31.7 & 25.9 & 5.8\\
        2.0m & $5.0^{\circ}$ & 32.3 & 25.4 & 6.9\\
        \hline
    \end{tabular}
\end{table}

Specifically, we evaluate performance under translation errors with $\sigma \in$ \{0.0, 0.5, 1.0, 2.0\} meters and rotation errors with $\sigma \in$ \{0.0$^\circ$, 2.5$^\circ$, 5.0$^\circ$\}. 
Sensitivity to pose noise is quantified by injecting these 
perturbations at training and test time.

Results in Tab.~\ref{table:localization_error} show performance at test time, with the performance drop from the baseline in parenthesis.
While mapping performance does go down, even with
error ranges up to 1m across all orientation errors, performance remains close to or above SOTA.
Error distributions with $\sigma=$ 2m perform slightly worse.



In Tab.~\ref{table:comparison_with_non-robust}, we evaluate “robust” models (trained with localization noise) against the “non-robust” model (no noise in training), testing both under the same seed-fixed perturbations for fairness and reproducibility. Robust training consistently outperforms the non-robust models under every non-zero perturbation.
The average gain of robust models across noisy conditions is \textbf{+3.86 mAP} (maximum +6.9). In terms of decay from the clean baseline, robust models cut the mean loss from 14.2 to 10.3 points and reduced the worst-case drop from -19.4 to -13.1 points. In practice, even small, realistic GPS drifts can cause steep collapses for a non-robust model, whereas the robust model degrades gracefully, preserving much of its clean performance.
In practice, even small, realistic GPS drifts can cause steep collapses for a non-robust model, whereas the robust model degrades gracefully, preserving much of its clean performance.

\subsection{Qualitative Evaluation}

\begin{figure*}[h]
    \centering
    \includegraphics[width=1.0\textwidth]{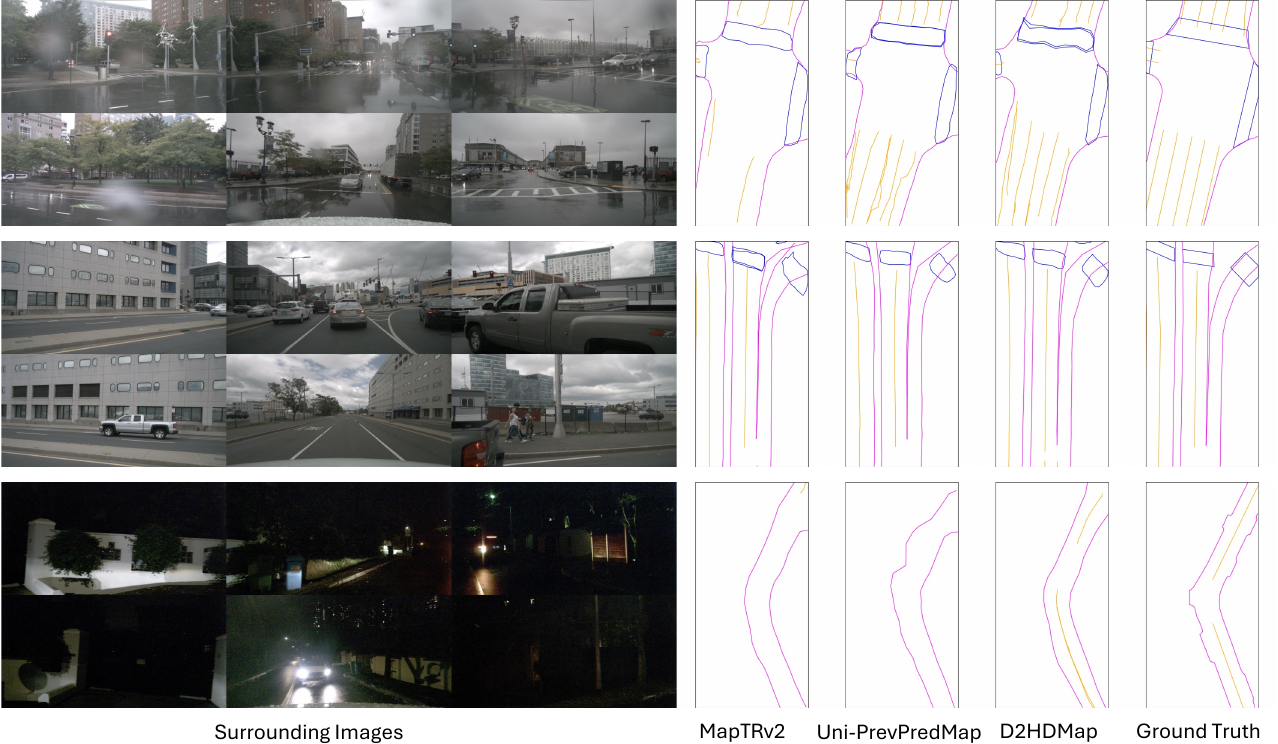}   
    \caption{Qualitative comparison of MapTRv2, Uni-PrevPredMap, and the proposed \ourAcronym{} across three distinct driving scenarios. Each row corresponds to a different scene: (top) rainy scene, (middle) daytime with occlusion, and (bottom) nighttime. Magenta, orange, and blue lines denote road boundaries, lane dividers, and pedestrian crossings, respectively. The visualization highlights each method’s ability to handle varying lighting and occlusion conditions.}
    \label{fig:visualization}
\end{figure*}

Fig.~\ref{fig:visualization} qualitatively compares inference results of three different models, MapTRv2 \cite{liao2025maptrv2}, Uni-PrevPredMap \cite{peng2025uni}, and \ourAcronym{}, in three challenging scenarios including rainy, daytime with occlusion, and low-visibility nighttime condition. To simulate real operation, we inject a seed-fixed localization error at inference of 0.5m translation and $2.5^{\circ}$ rotation. This misregisters any exterinal prior relative to the ego frame, and therefore affects the prior-consuming methods (Uni-PrevPredMap \cite{peng2025uni}, \ourAcronym). By contrast, the priors-free model, MapTRv2 \cite{liao2025maptrv2}, operates purely in the ego frame and is thus unaffected by the injected pose noise.

Across all three scenarios, \ourAcronym{} produces predictions of road structures that most closely align with ground truth even under unwelcoming scenarios with localization error. Uni-PrevPredMap \cite{peng2025uni} shows visible markings consistent with prior misalignment such as small lateral offset, kinks, and road boundary distortion, illustrating its higher sensitivity to localization error. MapTRv2 \cite{liao2025maptrv2} avoids localization error but lacks the structural anchor provided by a prior, yielding incomplete dividers, especially under poor visibility. The difference is most noticeable at night, where lane markings are not clearly visible: \ourAcronym{} correctly reconstructs the lane dividers and the geometry of road boundary. MapTRv2 and Uni-PrevPredMap miss the divider, and Uni-PrevPredMap exhibits kinks and distortion of the boundary due to injected pose noise. These trends are consistent with our quantitative results: noise-aware training makes \ourAcronym{} robust to realistic localization noise, and the driveline prior provides particularly strong cues for lane-aligned classes (dividers, boundaries).

%% file: 5_conclusion.tex
\section{Conclusion}

We introduced \ourAcronym, an online mapping system that leverages a driveline map prior for online vectorized HD map construction. The approach mitigates limitations of HD maps and purely sensor-based methods by using a non-visible, driveline prior. 
Models trained with driveline priors retain accuracy even when priors are missing (av2 geosplit -1.6 mAP vs. -7.1 / -8.2 for HD-map priors), demonstrating improved generalization and resilience to occlusion and appearance changes. 
Noise-aware training further enhances robustness, reducing localization drift impact (+3.86 mAP under noise, worst-case drop from -19.4 to -13.1 points)
Together with new state-of-the-art results on geographically disjoint splits, D2HDMap provides a scalable recipe for reliable online mapping: use small, widely available priors to constrain topology and train explicitly for pose noise. 
While driveline priors improve stability, large angular misalignment still causes minor degradation. Future work will explore additional lightweight priors to better capture crosswalks without reintroducing heavy HD-map dependence.



%% file: mapping.bib
@InProceedings{He_2016_CVPR,
author = {He, Kaiming and Zhang, Xiangyu and Ren, Shaoqing and Sun, Jian},
title = {Deep Residual Learning for Image Recognition},
booktitle = {Proceedings of the IEEE Conference on Computer Vision and Pattern Recognition (CVPR)},
month = {June},
year = {2016}
}

@article{oquab2023dinov2,
  title={Dinov2: Learning robust visual features without supervision},
  author={Oquab, Maxime and Darcet, Timoth{\'e}e and Moutakanni, Th{\'e}o and Vo, Huy and Szafraniec, Marc and Khalidov, Vasil and Fernandez, Pierre and Haziza, Daniel and Massa, Francisco and El-Nouby, Alaaeldin and others},
  journal={arXiv preprint arXiv:2304.07193},
  year={2023}
}

@article{huang2022bevpoolv2,
  title={Bevpoolv2: A cutting-edge implementation of bevdet toward deployment},
  author={Huang, Junjie and Huang, Guan},
  journal={arXiv preprint arXiv:2211.17111},
  year={2022}
}

@article{elghazaly2023high,
  title={High-definition maps: Comprehensive survey, challenges, and future perspectives},
  author={Elghazaly, Gamal and Frank, Rapha{\"e}l and Harvey, Scott and Safko, Stefan},
  journal={IEEE Open Journal of Intelligent Transportation Systems},
  volume={4},
  pages={527--550},
  year={2023},
  publisher={IEEE}
}

@misc{OpenStreetMap,
   author = {{OpenStreetMap contributors}},
   title = {{Planet dump retrieved from https://planet.osm.org }},
   howpublished = "\url{ https://www.openstreetmap.org }",
   year = {2017},
 }

@inproceedings{chen2024maptracker,
  title={Maptracker: Tracking with strided memory fusion for consistent vector hd mapping},
  author={Chen, Jiacheng and Wu, Yuefan and Tan, Jiaqi and Ma, Hang and Furukawa, Yasutaka},
  booktitle={European Conference on Computer Vision},
  pages={90--107},
  year={2024},
  organization={Springer}
}

@inproceedings{caesar2020nuscenes,
  title={nuscenes: A multimodal dataset for autonomous driving},
  author={Caesar, Holger and Bankiti, Varun and Lang, Alex H and Vora, Sourabh and Liong, Venice Erin and Xu, Qiang and Krishnan, Anush and Pan, Yu and Baldan, Giancarlo and Beijbom, Oscar},
  booktitle={Proceedings of the IEEE/CVF conference on computer vision and pattern recognition},
  pages={11621--11631},
  year={2020}
}

@article{wilson2023argoverse,
  title={Argoverse 2: Next generation datasets for self-driving perception and forecasting},
  author={Wilson, Benjamin and Qi, William and Agarwal, Tanmay and Lambert, John and Singh, Jagjeet and Khandelwal, Siddhesh and Pan, Bowen and Kumar, Ratnesh and Hartnett, Andrew and Pontes, Jhony Kaesemodel and others},
  journal={arXiv preprint arXiv:2301.00493},
  year={2023}
}

@misc{luo2023augmentinglaneperceptiontopology,
      title={Augmenting Lane Perception and Topology Understanding with Standard Definition Navigation Maps}, 
      author={Katie Z Luo and Xinshuo Weng and Yan Wang and Shuang Wu and Jie Li and Kilian Q Weinberger and Yue Wang and Marco Pavone},
      year={2023},
      eprint={2311.04079},
      archivePrefix={arXiv},
      primaryClass={cs.CV},
      url={https://arxiv.org/abs/2311.04079}, 
}

@inproceedings{li2022hdmapnet,
  title={Hdmapnet: An online hd map construction and evaluation framework},
  author={Li, Qi and Wang, Yue and Wang, Yilun and Zhao, Hang},
  booktitle={2022 International Conference on Robotics and Automation (ICRA)},
  pages={4628--4634},
  year={2022},
  organization={IEEE}
}

@inproceedings{liu2023vectormapnet,
  title={Vectormapnet: End-to-end vectorized hd map learning},
  author={Liu, Yicheng and Yuan, Tianyuan and Wang, Yue and Wang, Yilun and Zhao, Hang},
  booktitle={International Conference on Machine Learning},
  pages={22352--22369},
  year={2023},
  organization={PMLR}
}

@inproceedings{xiong2023neural,
  title={Neural map prior for autonomous driving},
  author={Xiong, Xuan and Liu, Yicheng and Yuan, Tianyuan and Wang, Yue and Wang, Yilun and Zhao, Hang},
  booktitle={Proceedings of the IEEE/CVF Conference on Computer Vision and Pattern Recognition},
  pages={17535--17544},
  year={2023}
}

@misc{tumu2025usinglanguageroadmanuals,
      title={Using Language and Road Manuals to Inform Map Reconstruction for Autonomous Driving}, 
      author={Akshar Tumu and Henrik I. Christensen and Marcell Vazquez-Chanlatte and Chikao Tsuchiya and Dhaval Bhanderi},
      year={2025},
      eprint={2506.10317},
      archivePrefix={arXiv},
      primaryClass={cs.RO},
      url={https://arxiv.org/abs/2506.10317}, 
}

@inproceedings{yuan2024streammapnet,
  title={Streammapnet: Streaming mapping network for vectorized online hd map construction},
  author={Yuan, Tianyuan and Liu, Yicheng and Wang, Yue and Wang, Yilun and Zhao, Hang},
  booktitle={Proceedings of the IEEE/CVF Winter Conference on Applications of Computer Vision},
  pages={7356--7365},
  year={2024}
}

@inproceedings{liao2023maptr,
title={Map{TR}: Structured Modeling and Learning for Online Vectorized {HD} Map Construction},
author={Bencheng Liao and Shaoyu Chen and Xinggang Wang and Tianheng Cheng and Qian Zhang and Wenyu Liu and Chang Huang},
booktitle={The Eleventh International Conference on Learning Representations },
year={2023},
url={https://arxiv.org/abs/2208.14437}
}

@article{liao2025maptrv2,
  title={Maptrv2: An end-to-end framework for online vectorized hd map construction},
  author={Liao, Bencheng and Chen, Shaoyu and Zhang, Yunchi and Jiang, Bo and Zhang, Qian and Liu, Wenyu and Huang, Chang and Wang, Xinggang},
  journal={International Journal of Computer Vision},
  volume={133},
  number={3},
  pages={1352--1374},
  year={2025},
  publisher={Springer}
}

@inproceedings{zhang2024enhancingvectorized,
  title={Enhancing vectorized map perception with historical rasterized maps},
  author={Zhang, Xiaoyu and Liu, Guangwei and Liu, Zihao and Xu, Ningyi and Liu, Yunhui and Zhao, Ji},
  booktitle={European Conference on Computer Vision},
  pages={422--439},
  year={2024},
  organization={Springer}
}

@inproceedings{peng2025prevpredmap,
  title={Prevpredmap: Exploring temporal modeling with previous predictions for online vectorized hd map construction},
  author={Peng, Nan and Zhou, Xun and Wang, Mingming and Yang, Xiaojun and Chen, Songming and Chen, Guisong},
  booktitle={2025 IEEE/CVF Winter Conference on Applications of Computer Vision (WACV)},
  pages={8134--8143},
  year={2025},
  organization={IEEE}
}

@article{peng2025uni,
  title={Uni-PrevPredMap: Extending PrevPredMap to a Unified Framework of Prior-Informed Modeling for Online Vectorized HD Map Construction},
  author={Peng, Nan and Zhou, Xun and Wang, Mingming and Chen, Guisong and Xu, Wenqi},
  journal={arXiv preprint arXiv:2504.06647},
  year={2025}
}

@inproceedings{Bateman2024hdmapprior,
  author    = {Bateman, Samuel M. and Xu, Ning and Zhao, H. Charles and Shalon, Yae Ben and Gong, Vince and Long, Greg and Maddern, Will},
  title     = {Exploring Real World Map Change Generation for Prior-informed HD Map Prediction Models},
  year      = {2024},
  booktitle = {2024 IEEE/CVF Conference on Computer Vision and Pattern Recognition Workshops (CVPRW)}
}

@inproceedings{sun2025mind,
  title={Mind the Map! Accounting for Existing Maps When Estimating Online HDMaps from Sensors},
  author={Sun, Remy and Yang, Li and Lingrand, Diane and Precioso, Fr{\'e}d{\'e}ric},
  booktitle={2025 IEEE/CVF Winter Conference on Applications of Computer Vision (WACV)},
  pages={1671--1681},
  year={2025},
  organization={IEEE}
}

@inproceedings{zhang2024ucsd,
  title={Enhancing online road network perception and reasoning with standard definition maps},
  author={Zhang, Hengyuan and Paz, David and Guo, Yuliang and Das, Arun and Huang, Xinyu and Haug, Karsten and Christensen, Henrik I and Ren, Liu},
  booktitle={2024 IEEE/RSJ International Conference on Intelligent Robots and Systems (IROS)},
  pages={1086--1093},
  year={2024},
  organization={IEEE}
}

@article{xia2025ldmapnet,
  title={LDMapNet-U: An End-to-End System for City-Scale Lane-Level Map Updating},
  author={Xia, Deguo and Zhang, Weiming and Liu, Xiyan and Zhang, Wei and Gong, Chenting and Tan, Xiao and Huang, Jizhou and Yang, Mengmeng and Yang, Diange},
  journal={arXiv preprint arXiv:2501.02763},
  year={2025}
}

@article{diwanji2025sd++,
  title={SD++: Enhancing Standard Definition Maps by Incorporating Road Knowledge using LLMs},
  author={Diwanji, Hitvarth and Liao, Jing-Yan and Tumu, Akshar and Christensen, Henrik I and Vazquez-Chanlatte, Marcell and Tsuchiya, Chikao},
  journal={arXiv preprint arXiv:2502.02773},
  year={2025}
}

@article{immel2025sdtagnet,
  title={SDTagNet: Leveraging Text-Annotated Navigation Maps for Online HD Map Construction},
  author={Immel, Fabian and Pauls, Jan-Hendrik and Fehler, Richard and Bieder, Frank and Merkert, Jonas and Stiller, Christoph},
  journal={arXiv preprint arXiv:2506.08997},
  year={2025}
}

@article{ye2025smart,
  title={Smart: Advancing scalable map priors for driving topology reasoning},
  author={Ye, Junjie and Paz, David and Zhang, Hengyuan and Guo, Yuliang and Huang, Xinyu and Christensen, Henrik I and Wang, Yue and Ren, Liu},
  journal={arXiv preprint arXiv:2502.04329},
  year={2025}
}

@article{jiang2024p,
  title={P-mapnet: Far-seeing map generator enhanced by both sdmap and hdmap priors},
  author={Jiang, Zhou and Zhu, Zhenxin and Li, Pengfei and Gao, Huan-ang and Yuan, Tianyuan and Shi, Yongliang and Zhao, Hang and Zhao, Hao},
  journal={IEEE Robotics and Automation Letters},
  year={2024},
  publisher={IEEE}
}

@article{yang2025histrackmap,
  title={Histrackmap: Global vectorized high-definition map construction via history map tracking},
  author={Yang, Jing and Yang, Sen and Tan, Xiao and Wang, Hanli},
  journal={arXiv preprint arXiv:2503.07168},
  year={2025}
}

@article{jia2025enhancing,
  title={Enhancing Lane Segment Perception and Topology Reasoning with Crowdsourcing Trajectory Priors},
  author={Jia, Peijin and Luo, Ziang and Wen, Tuopu and Yang, Mengmeng and Jiang, Kun and Cui, Le and Yang, Diange},
  journal={IEEE Robotics and Automation Letters},
  year={2025},
  publisher={IEEE}
}

@inproceedings{hubbertz2025inferring,
  title={Inferring Driving Maps by Deep Learning-based Trail Map Extraction},
  author={Hubbertz, Michael and Colling, Pascal and Han, Qi and Meisen, Tobias},
  booktitle={Proceedings of the Computer Vision and Pattern Recognition Conference},
  pages={2425--2434},
  year={2025}
}

@article{zhu2023nemo,
  title={Nemo: Neural map growing system for spatiotemporal fusion in bird's-eye-view and bdd-map benchmark},
  author={Zhu, Xi and Cao, Xiya and Dong, Zhiwei and Zhou, Caifa and Liu, Qiangbo and Li, Wei and Wang, Yongliang},
  journal={arXiv preprint arXiv:2306.04540},
  year={2023}
}

@article{monninger2025mapdiffusion,
  title={MapDiffusion: Generative Diffusion for Vectorized Online HD Map Construction and Uncertainty Estimation in Autonomous Driving},
  author={Monninger, Thomas and Zhang, Zihan and Mo, Zhipeng and Anwar, Md Zafar and Staab, Steffen and Ding, Sihao},
  journal={arXiv preprint arXiv:2507.21423},
  year={2025}
}

@article{Reid_2019,
   title={Localization Requirements for Autonomous Vehicles},
   volume={2},
   ISSN={2574-075X},
   url={http://dx.doi.org/10.4271/12-02-03-0012},
   DOI={10.4271/12-02-03-0012},
   number={3},
   journal={SAE International Journal of Connected and Automated Vehicles},
   publisher={SAE International},
   author={Reid, Tyler G.R. and Houts, Sarah E. and Cammarata, Robert and Mills, Graham and Agarwal, Siddharth and Vora, Ankit and Pandey, Gaurav},
   year={2019},
   month=sep }

@inproceedings{lilja2024localization,
  title={Localization is all you evaluate: Data leakage in online mapping datasets and how to fix it},
  author={Lilja, Adam and Fu, Junsheng and Stenborg, Erik and Hammarstrand, Lars},
  booktitle={Proceedings of the IEEE/CVF Conference on Computer Vision and Pattern Recognition},
  pages={22150--22159},
  year={2024}
}
